\documentclass[runningheads]{llncs}

\usepackage{eccv}

\usepackage{eccvabbrv}

\usepackage{graphicx}
\usepackage{booktabs}
\usepackage{ulem}

\usepackage[accsupp]{axessibility}  %

\usepackage{amssymb}%
\usepackage{pifont}%

\newcommand{\mdred}{\color[HTML]{F44336}}
\newcommand{\mdblue}{\color[HTML]{2196F3}}
\newcommand{\mdgreen}{\color[HTML]{4CAF50}}

\newcommand{\image}{\mathbf{I}}

\usepackage{array}
\newcolumntype{H}{>{\setbox0=\hbox\bgroup}c<{\egroup}@{}}
\newcolumntype{L}[1]{>{\raggedright\let\newline\\\arraybackslash\hspace{0pt}}m{#1}}
\newcolumntype{C}[1]{>{\centering\let\newline\\\arraybackslash\hspace{0pt}}m{#1}}
\newcolumntype{R}[1]{>{\raggedleft\let\newline\\\arraybackslash\hspace{0pt}}m{#1}}

\usepackage{acronym}
\acrodef{VO}[VO]{Visual Odometry}
\acrodef{SLAM}[SLAM]{Simultaneous Localization and Mapping}
\acrodef{VSLAM}[VSLAM]{Visual SLAM}
\acrodef{MV}[MV]{motion vectors}
\acrodef{MB}[MB]{Macro Block}
\acrodef{MSD}[MSD]{Monado SLAM dataset}

\def\estimated{\mathbf{T}}
\def\estimatedT{\mathbf{t}}
\def\estimatedR{\mathbf{R}}
\def\groundtruth{\tilde{\mathbf{T}}}
\def\groundtruthT{\tilde{\mathbf{t}}}
\def\groundtruthR{\tilde{\mathbf{R}}}

\usepackage{listings}

\makeatletter
\AtBeginDocument
 {
   \def\ltx@label#1{\cref@label{#1}}%
   \def\label@in@display@noarg#1{\cref@old@label@in@display{#1}}%
\def\label@in@mmeasure@noarg#1{%
    \begingroup%
      \measuring@false%
      \cref@old@label@in@display{#1}%
    \endgroup}%
 } %
\makeatother

\usepackage{enumitem}

\usepackage{hyperref}

\usepackage{orcidlink}

\newread\imgstream
\immediate\openin\imgstream=imagedata.in
\makeatletter
\def\new@kvginclip#1{}
\def\new@kvgintrim#1{}
\let\old@kvginclip\KV@Gin@clip
\let\old@kvgintrim\KV@Gin@trim
\let\oldincludegraphics\includegraphics
\providecommand{\includegraphics}{}
\renewcommand{\includegraphics}[2][]{%
  \immediate\read\imgstream to \src
  \immediate\read\imgstream to \removecrop
  \ifnum\removecrop=1
      \let\KV@Gin@clip\new@kvginclip
      \let\KV@Gin@trim\new@kvgintrim
  \fi
  \oldincludegraphics[#1]{\src}%
  \let\KV@Gin@clip\old@kvginclip
  \let\KV@Gin@trim\old@kvgintrim}
\makeatother

\begin{document}

\title{VOCA: Visual Odometry with Codec Awareness}

\author{Nouri Alexander Hilscher\thanks{equal contribution}\inst{1}\orcidlink{0009-0001-0293-8153} \hspace{0.4cm}
Mateo de Mayo$^{*}$\inst{1,2}\orcidlink{0009-0003-0729-3838} \hspace{0.4cm}
Dominik Muhle\inst{1,2}\orcidlink{0009-0007-2606-6541} \\
Christoph Otten genannt Hermes\inst{1} \hspace{0.6cm}
Daniel Cremers\inst{1,2}\orcidlink{0000-0002-3079-7984}
}

\authorrunning{N.~Hilscher et al.}

\institute{Technical University of Munich, Munich, Germany \and
Munich Center for Machine Learning, Munich, Germany
}

\maketitle

\begin{figure}
  \centering
  \includegraphics[width=\linewidth]{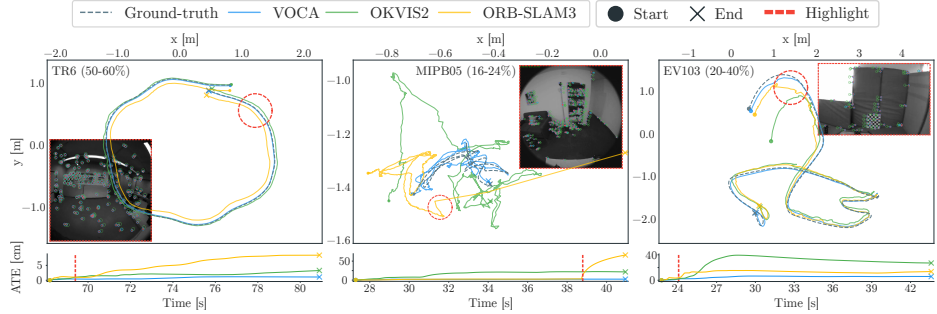}
  \vspace{-1em}
  \caption{\textbf{Visual Odometry on Compressed Videos.} We present VOCA, a novel Visual Odometry system that produces smoother, more stable trajectories than descriptor-based systems such as ORB-SLAM3 and OKVIS2, thanks to its codec-aware sparse optical-flow frontend. Our system enables Visual Odometry on data compressed by up to $100\times$. We visualize challenging segments with dashed-red markers, sampled from three different datasets. In the zoomed-in views, we show the \textbf{\mdgreen previous pixel} location that would be the prior in regular VO systems, the \textbf{\mdblue motion-vector} prior that we introduce, and the optical flow \textbf{\mdred solution}. In most cases, motion-vector priors reduce the initialization distance to the solution to just a few pixels.}
  \label{fig:trajectories}
\end{figure}

\vspace{-2.5em}
\begin{abstract}
  Camera pose estimation from image streams is a critical component of spatial
  world models that integrate perception into planning and decision-making. Nearly
  all \acf{VO} and \acf{SLAM} systems have focused on datasets containing raw,
  uncompressed videos. Many working systems instead use ubiquitous hardware
  units to efficiently compress and decode video streams, saving orders of
  magnitude in storage and bandwidth. However, this lossy compression introduces
  visual artifacts that hinder the performance of traditional tracking systems. We present VOCA, a causal stereo visual-odometry method that exploits codec
  information to improve tracking performance. We achieve state-of-the-art
  performance on causal \ac{VO} for relative trajectory error, efficiency, and
  absolute trajectory error on compressed streams. This work highlights the
  potential of leveraging widely available video codec information for vision tasks.
  \keywords{Visual Odometry \and KLT-Tracking \and Compressed Videos}
\end{abstract}

\section{Introduction}
\label{sec:intro}

Spatial computing systems make their way into our lives in a variety of forms,
from mixed reality devices to autonomous systems such as robotic assistants,
drones, and self-driving cars. Cameras have proven to be a rich and cost-effective sensor,
driven in part by the ubiquity of mobile and embedded vision in consumer devices \cite{banerjeeHOT3DHandObject2025,ungureanuHoloLens2Research2020a}.
Current trends in computer vision, learning, and robotics show increasing overlap in research, with no sign of slowing \cite{chen2025vidbot, Ji2025RoboBrainAU, qian3DMVP3DMultiview}.
Furthermore, immense potential is expected from learning-based models that leverage the vast amount of internet videos captured by these sensors for training \cite{bahlAffordancesHumanVideos2023, chen2025vidbot}.
In this context, tracking camera poses has become imperative for many applications and research \cite{chi2026c3po, han2024boosting, hayler2024s4c, wimbauer2023behind, gross2026ipformer}. There
cannot be action and planning in 3D without first understanding the current
location of a moving agent \cite{hartleyMultipleViewGeometry2004}.

The high potential for cameras comes at the cost of large volumes of data. A raw image stream from a common setup of stereo monochrome cameras (8-bit pixels) with a
resolution of 640 $\times$ 480 at 30 frames per second produces more than a gigabyte
of data per minute. Since bandwidth and memory are expensive \cite{zhuMEETMemoryEfficientTemporal2025}, practical systems rely on compression to transmit camera streams efficiently under the compress-then-analyze design paradigm \cite{hoferH264CompressThenAnalyzeTransmission2023}. Years of software
and hardware development have gone into accelerating common encoders such as H.264,
AV1, and VP9 \cite{h264_overview_paper, AV1_overview_paper, VP9_overview_paper}. Video codecs are now ubiquitous when working with
camera streams \cite{h264_domination_2024}.

Over the last 20 years, a variety of approaches, ranging from purely classical
algorithms \cite{ORB_SLAM_Mur-Artal2015, ORB_SLAM2_Mur-Artal2017,
ORB_SLAM3_campos2021, DSO_Engel2016, Basalt_Usenko2020, leutenegger2022okvis2,
ruckert2021snake, schonberger2016structure, geneva2020openvins} to various
learning-based systems \cite{wang2025vggt, keetha2025mapanything,
wimbauer2025anycam, wang2025pi, lin2025depth, smith2025flowmap, muhle2023learning} have been
proposed in the literature for egocentric tracking. However, most \ac{VO} and
\ac{VSLAM} systems focus solely on raw images, with datasets and benchmarks in
the field considering only fully uncompressed images
\cite{de2025monado,banerjeeHOT3DHandObject2025}. As we will show, common video encoders like
H.264 can reduce the size of sequences by orders of magnitude. Unfortunately,
this compression process is lossy and introduces visual artifacts that
significantly degrade the performance of these state-of-the-art tracking systems
\cite{turner2023mov} and other vision tasks \cite{reich2024perspective}. This
has been an often overlooked challenge in the field of \ac{VO} and \ac{VSLAM}, with
only a few works discussing the topic \cite{zouein2025leveraging,
hoferH264CompressThenAnalyzeTransmission2023} and, to the best of our knowledge,
only one, MoV-SLAM \cite{turner2023mov}, attempting to actively improve tracking
through codec information.

In this work, we show that a significant amount of the tracking performance loss due to compression can be recovered by properly exploiting the information available
in the encoded data. In some cases, certain appearance changes
introduced by compression can make feature tracking more stable (e.g., lower
pixel noise since it is difficult to
encode). In addition, the reduced bandwidth and memory requirements of compressed streams can improve runtime performance, since decoding can, in many cases, be more efficient than repeatedly accessing the much larger raw image sequences.

Following these observations, we present VOCA, a \textit{causal} stereo \ac{VO} method that
leverages video codec information for tracking. VOCA is built on top of the
popular odometry system Basalt \cite{Basalt_Usenko2020} with modifications from \cite{de2025monado}. It produces causal estimates, meaning it uses only past information for tracking, making it suitable for real-time use \cite{leutenegger2022okvis2}. Our method
outperforms Basalt on $\sim100\times$ compressed sequences. Furthermore, VOCA achieves state-of-the-art performance for causal stereo \ac{VO} when compared to other top-scoring
systems (see \cref{fig:trajectories}). The contributions of our work include:
\begin{itemize}
    \item VOCA, a causal stereo \ac{VO} method that can run on compressed videos and
    leverage its codec information for greater accuracy and efficiency.
    \item An extensive benchmark showing that VOCA achieves state-of-the-art
    performance, even in settings with heavy compression ($\sim100\times$).
    \item An ablation study that isolates the contribution of each component and provides insights for future methods for tracking encoded videos.
    \item A performance degradation analysis, showing tracking degradation for
    top-scoring systems under different compression settings
\end{itemize}
See VOCA's project page at \texttt{\mdblue \href{https://tum-vision.github.io/voca}{https://tum-vision.github.io/voca}}.

\begin{figure}[tb]
  \centering
  \includegraphics[width=\linewidth]{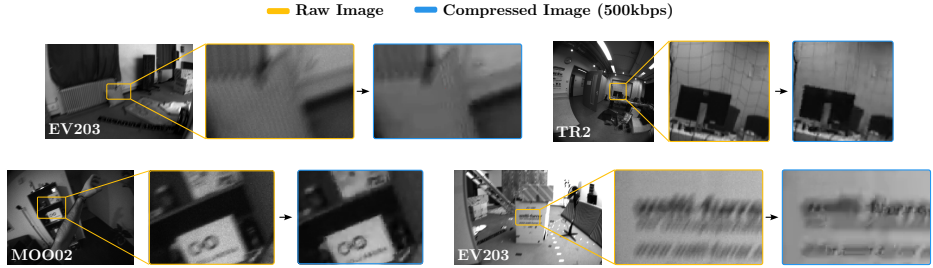}
  \caption{Video encoding can introduce artifacts
    that violate the photometric-constancy assumption used by most tracking algorithms. Examples from datasets used in this work: EV203 (from EuRoC \cite{burri2016euroc}) shows blurred details, TR2 (from TUM-VI room \cite{schubert2018tum}) exhibits reduced contrast in the thin net, and MOO02 (from MSD \cite{de2025monado}) contains geometrically jagged edges/textures. See \cite{linPEA265PerceptualAssessment2019} for additional artifacts.}
  \label{fig:artifacts}
\end{figure}
\section{Related Work}
\label{sec:related_work}

The focus of this paper is \ac{VO} for compressed images. Our discussion of related work will include an overview of the topics \textit{\ac{VO}} and  \textit{\ac{VSLAM}}, with a focus on the most closely related work MoV-SLAM \cite{turner2023mov}. We include approaches that also use inertial measurements, although our work is primarily concerned with improvements in front-end tracking. Therefore, we will introduce feature tracking separately to discuss its challenges under video compression. We will also give a brief introduction to video compression.

\subsection{Feature Tracking}
Feature-based \ac{VO} and \ac{SLAM} approaches require tracking feature positions across multiple images to construct a (windowed) bundle adjustment optimization problem. To establish these correspondences, feature positions are extracted and descriptors of these feature positions are matched between images.

For classical \ac{VO}, popular descriptors are SIFT/SURF \cite{lowe2004distinctive, bay2006surf}, ORB \cite{ORB_Rublee2011}, and BRISK \cite{leutenegger2011brisk}.  SuperPoint \cite{detone2018superpoint} presents a learned alternative. For compressed videos, MoV-SLAM \cite{turner2023mov} introduced EXPRESS features based on the \acp{MB} of the H.264 video encoder. Each \ac{MB} represents a potential feature, classified as \textit{distinctive} or not, based on a linear-time operation along the patch diagonals. The features are scale-invariant but not rotation-invariant. They are matched using a binary descriptor distance, similar to ORB features.

Tracking feature points presents an alternative to feature matching, usually based on optical flow, such as KLT-Tracking \cite{KLT_Shi1994, KLT_Lukas1981, KLT_Tomasi1991}. The reliance on pixel intensities renders optical-flow-based approaches sensitive to image artifacts introduced by compression, as the photo constancy assumption is violated. In this work, we exploit information from video compression to reduce the impact of these artifacts on the feature tracking, improving its robustness.

\subsection{\acl{VO}}
Classical \ac{VO} approaches can be broadly categorized into feature-based and direct methods. Whereas feature-based methods such as PTAM \cite{klein2007parallel}, MRO \cite{MRO_Chng2020}, ROBA \cite{ROBA_Lee2020}, PNEC \cite{muhle2022probabilistic}, and Basalt \cite{Basalt_Usenko2020} use the aforementioned feature positions, direct methods such as DSO \cite{DSO_Engel2016} and DM-VIO \cite{von2022dm} operate directly on pixel intensities. Because they rely on the photo constancy assumption, direct methods are even more sensitive to image artifacts than sparse feature-tracking methods. More recently, purely learning-based approaches have enabled \ac{VO} on uncalibrated videos including VGGT \cite{wang2025vggt}, PI3 \cite{wang2025pi}, MapAnything \cite{keetha2025mapanything}, AnyCam \cite{wimbauer2025anycam}, AnyMap \cite{cin2025anymap}, and DepthAnything3 \cite{lin2025depth}. Due to their large model size and scalability constraints for long videos, they are not yet viable for deployment on constrained hardware. Hybrid approaches like LEAP-VO \cite{chen2024leap} and Splat-SLAM \cite{sandstrom2025splat} bridge the gap somewhat. We build our work on Basalt \cite{Basalt_Usenko2020} as it is among the few \ac{VO} systems that use KLT-Tracking.

\subsection{Visual \acl{SLAM}}
Despite its rigorous mathematical definition in \cite{stachniss2016simultaneous}, we follow the commonly accepted approach in which \ac{SLAM} extends \ac{VO} with concepts such as loop closure \cite{carlone2025slam}. Similar to \ac{VO}, a family of direct \ac{SLAM} methods exists around LSD-SLAM \cite{engel2014lsd}. ORB-SLAM 1, 2, and 3 \cite{ORB_SLAM_Mur-Artal2015, ORB_SLAM3_campos2021, ORB_SLAM2_Mur-Artal2017} build a system around matching ORB features and loop detection using Bag-of-Words. OKVIS2 \cite{leutenegger2022okvis2} employs BRISK features and is optimized for drone usage. Snake-SLAM \cite{ruckert2021snake} decouples front-end processing and back-end optimization, which is further split into IMU state estimation and bundle adjustment to improve efficiency. Similar to \ac{VO}, end-to-end learning approaches \cite{teed2021droid, murai2025mast3r, deng2025vggt} have been introduced in recent years. More closely related to our approach is MoV-SLAM \cite{turner2023mov}. MoV-SLAM leverages the EXPRESS features discussed previously for front-end correspondence estimation with optional stereo matching using Lucas-Kanade optical flow. The back-end is based on PnP pose estimation using MAGSAC++. Despite its name, MoV-SLAM does not use loop closure for drift correction.

\subsection{Video Compression}
Early digital video coding was first standardized for low-bitrate transmission (e.g., ITU-T H.261 \cite{h261_and_history}), and later for broadcast and storage (MPEG-2~\cite{h262_website}), which combined block transforms with motion-compensated inter-frame prediction \cite{387092}. With the rise of videos on the internet, H.264/AVC~\cite{h264_overview_paper} became the most widely adopted codec by major platforms \cite{h264_domination_2011, h264_domination_2015, h264_domination_2024}.
Modern video codecs (e.g., MPEG-2/H.262, H.264/AVC, H.265/HEVC, VP9, AV1, and H.266/VVC \cite{h262_website, h264_overview_paper, h265_overview_paper, VP9_overview_paper, AV1_overview_paper, H266_overview_paper}) largely follow a hybrid block-based architecture: Each frame is partitioned into blocks, and for each block the codec forms a predictor via intra- or inter-prediction.
The former leverages previously reconstructed neighboring blocks within the same frame, while the latter uses motion-compensated prediction from one or more previously decoded reference frames, either causally or not, depending on encoder settings. The block is then transformed, quantized, and entropy-coded, typically with in-loop filtering to improve reference quality.

For inter-frame prediction, the encoder selects one (or more) previously reconstructed reference frames and estimates per-block motion vectors that point to displaced regions in the reference; the decoder uses the corresponding displaced (and, if needed, sub-pixel interpolated) reference block as an initial prediction and reconstructs the final block by adding the decoded (inverse-quantized/inverse-transformed) residual. Although these per-block motion vectors are primarily optimized for rate–distortion efficiency in inter-frame coding, prior work shows that they roughly resemble optical flow (see \cref{fig:codec}) and can be reused as a  warm-start/initialization for optical flow estimation pipelines \cite{cvpr_optical_flow_action_recognition, zouein2025leveraging, av1_motion_vector_for_flow, mvflow}.

Despite newer codecs improving compression efficiency, H.264/AVC remains the most widely adopted codec on commodity devices \cite{h264_domination_2024}. Moreover, recent fixed-function encoder benchmarks show that H.264 encoding throughput often matches or even exceeds that of newer codecs \cite{evaluation_encoders}, owing to its mature hardware-accelerated implementations. Accordingly, we focus on H.264/AVC as our primary codec throughout the paper and refer the reader to the supplementary material for preliminary positive results of the method on the AV1 codec.
\section{Method}
We first explain why video compression degrades KLT-Tracking, and hence why systems like Basalt~\cite{Basalt_Usenko2020} struggle
under these conditions, then show how the compression itself can
initialize tracking to overcome these limitations. Throughout, we assume a causal
configuration in which encoding order matches capture order and
inter-frame prediction references only past frames. This matches both our target streaming setting
and all experiments (Sec.~\ref{sec:experiments}).
We begin with the basics of video compression and optical flow.
\label{sec:method}

\subsection{A Primer on Video Compression}
\label{sec:video_coding}
Let $\image_t:\Omega\to\mathbb{R}^C$ be the image at
time $t$ defined on the pixel lattice
$\Omega=\{1,\dots,H\}\times\{1,\dots,W\}$ with $C$ color channels. In video coding,
$\image_t$ is typically represented in a color space that separates the luminance, like YCbCr, and the
encoding pipeline operates on each component separately. For notational
simplicity and because the evaluated datasets are monochromatic, we describe the
process for a single scalar component
$Y_t:\Omega\to\mathbb{R}$.

Let $\{B_{k}\}_{k=1}^K$ be a frame-specific partition of $\Omega$ into $K$
rectangular blocks, indexed by $k$ alone since $t$ is fixed throughout. Each block
$B_k=\{m_k,\dots,m_k+H_k-1\}\times\{n_k,\dots,n_k+W_k-1\}
\subseteq\Omega$ has height $H_k$, width $W_k$, and a top-left anchor coordinate $(m_k,n_k)$. We define the block of component
values associated with $B_k$ as
$\mathbf{y}_{t,\,k} = Y_t(B_k) \in \mathbb{R}^{H_k \times W_k}$.

The encoder aims to represent each block
$\mathbf{y}_{t,k}$ as accurately as possible while minimizing the number of required bits. Within a frame, blocks are processed
in encoding order; when processing $B_k$, previously reconstructed
blocks $\hat{\mathbf{y}}_{t,1}, \dots, \hat{\mathbf{y}}_{t,k-1}$
within the current frame and reconstructed frames
$\hat{Y}_{t'}$, $t' < t$, are available. The encoder forms a
predicted block $\tilde{\mathbf{y}}_{t,k}$ either by
\textit{intra-frame} prediction from neighboring boundary samples
or by \textit{inter-frame} prediction, in which a reference frame
$\hat{Y}_{t'}$ and a displacement vector, often called a
\textit{motion vector},
$\mathbf{d}_k\in\mathbb{R}^2$, likewise indexed only by $k$, are selected by minimizing a
rate-distortion cost, yielding
$\tilde{\mathbf{y}}_{t,k} = \hat{Y}_{t'}(B_k + \mathbf{d}_k)$.
The prediction residual is then computed, transformed, and quantized as
\begin{equation}
    \mathbf{r}_{t,k} = \mathbf{y}_{t,k} -
    \tilde{\mathbf{y}}_{t,k}, \qquad
    \mathbf{X}_{t,k} =
    \mathcal{Q}\big(\mathcal{T}(\mathbf{r}_{t,k})\big).
\end{equation}
Since quantization is a lossy, many-to-one mapping, the decoder
can only recover an approximation
$\hat{\mathbf{r}}_{t,k} = \mathcal{T}^{-1}
(\mathcal{Q}^{-1}(\mathbf{X}_{t,k})) \neq \mathbf{r}_{t,k}$, which leads to artifacts as shown in Fig. \ref{fig:artifacts}.
The quantized
coefficients $\mathbf{X}_{t,k}$, together with the prediction
parameters (prediction mode, motion vector, and reference index), are
entropy coded into the bitstream. The choice of admissible
prediction types is governed at the frame level: within so-called
\textit{I-frames}, all blocks are restricted to intra-prediction,
making the frame independently decodable and serving as a
random-access point; within so-called \textit{P-frames}, each
block may choose between intra and inter prediction, and blocks
predicted in intra mode do not carry motion vectors. The large
majority of frames, however, are encoded as P-frames.

\begin{figure}[tb]
  \centering
  \includegraphics[width=\linewidth]{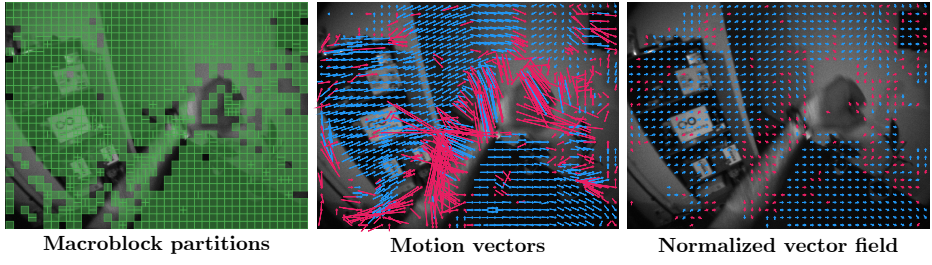}
  \caption{ \textbf{Encoded information.} Video codecs assign motion vectors to macroblock partitions and sub-partitions. They encode local block motion as a displacement to an intensity-matched block in a reference frame. The figure shows a frame with macroblock partitions (left), its motion vectors (center), and the normalized discrete vector field they induce (right). While this field is correlated with optical flow, it is considerably noisier: the encoder is free to match any region of the reference frame to improve compression, regardless of whether the resulting vector reflects true scene motion. Vectors likely deviating from optical flow are highlighted in red via median thresholding.}
  \label{fig:codec}
\end{figure}

In H.264 ~\cite{h264_overview_paper}, the blocks $B_k$ correspond to
so-called \textit{macroblocks} of fixed size $16\times 16$ or (sub)-partitions of macroblocks with sizes ranging from 16x8, 8x8, 4x8, 8x4, down to 4x4, each carrying its own
motion vector (see Fig. \ref{fig:codec}).

\subsection{From Video Compression to Lucas-Kanade Optical Flow}
We denote with $I_t: \Omega \rightarrow [0, 1]$ the gray-scale intensity of the image frame $\image_t$. The Lucas-Kanade \cite{KLT_Lukas1981} method assumes
\textit{brightness constancy} along the trajectory $\mathbf{p}(t)$
of a moving point in an image sequence $I(\mathbf{p},t)$,
\begin{equation}
I(\mathbf{p}(t),t) = \text{const}, \quad \forall t,
\end{equation}
which, via first-order linearization, yields the optical-flow
constraint
\begin{equation}
\frac{d}{dt} I(\mathbf{p}(t),t)
\;=\;
\nabla I(\mathbf{p},t)^\top \mathbf{v}(\mathbf{p},t)
+ \frac{\partial}{\partial t} I(\mathbf{p},t)
\;=\; 0,
\end{equation}
where $\mathbf{v}=(u,v)^\top$ denotes the (instantaneous)
optical flow and $I(\mathbf{p}, t) \equiv I_t(\mathbf{p})$.

Evaluated
between two consecutive frames (i.e., discretizing with
$\Delta t = 1$), the brightness constancy assumption takes the
form
\begin{equation}\label{eq:discrete_bc}
I_{t-1}(\tilde{\mathbf{p}}) \;\approx\;
I_{t}(\tilde{\mathbf{p}} + \mathbf{v})\,.
\end{equation}
Lucas-Kanade further assumes locally constant motion, i.e., $\mathbf{v}(\tilde{\mathbf{p}},t)\approx
\mathbf{v}(\mathbf{p},t)$ for all $\tilde{\mathbf{p}}$ in a
neighborhood $\mathcal{N}(\mathbf{p})$.

This local-constancy assumption mirrors the structure of
block-based inter-frame prediction in video coding: identifying
$\mathcal{N}(\mathbf{p})$ with the block $B_k$ containing
$\mathbf{p}$, the encoder assigns a motion vector
$\mathbf{d}_k$ that predicts $B_k$ by sampling a displaced
region from a reference frame, yielding the relation
\begin{equation}\label{eq:mv_approx}
{I}_{t}(\tilde{\mathbf{p}}) \;\approx\;
I_{t-1}(\tilde{\mathbf{p}}+\mathbf{d}_k),
\quad \forall \tilde{\mathbf{p}}\in B_k.
\end{equation}
Comparing with \cref{eq:discrete_bc}, since the motion
vector $\mathbf{d}_k$
points from the current frame back into the reference frame,
whereas $\mathbf{v}$ points forward in time, $-\mathbf{d}_k$ acts as a coarse, block-wise
approximation of sparse optical flow, i.e.,
$-\mathbf{d}_k \approx \mathbf{v}$.
However, as \cref{fig:codec} illustrates, this correspondence
can break down: since the encoder is free to reference any
region that improves compression, regardless of whether the
resulting vector reflects true scene motion, decoded motion
vectors may deviate substantially from optical flow,
particularly in areas with little texture, repetitive patterns, or strong motion blur. We therefore maintain a distinction between
the \textit{true optical flow} $\mathbf{v}$ and the decoded
\textit{motion vectors} $\mathbf{d}_k$ throughout.

\begin{figure}[t]
  \centering
  \includegraphics[width=0.95\linewidth]{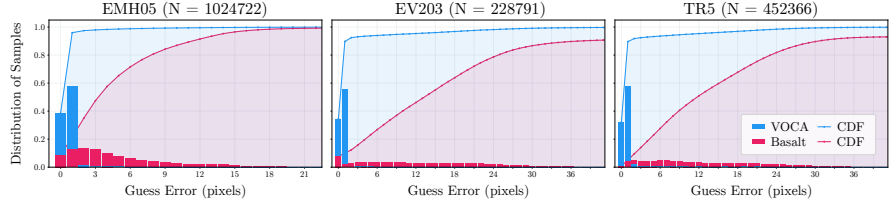}
  \caption{\textbf{Proximity to the minimum.} VOCA uses motion vectors as priors for optical flow, which significantly reduces the distance to the ground-truth solution, possibly improving convergence times and the likelihood that the initial state lies in the convergence basin of the non-linear problem. This figure shows the error distributions of pixel distances between the prior guesses (initialization) and the final tracked results.}
  \label{fig:guess_error}
\end{figure}
\subsection{Using Motion Vectors as Priors for KLT-Tracking}
While video-codec motion vectors provide purely translational
displacements per block, KLT-Tracking estimates correspondences
more accurately by optimizing over a richer motion model for
image patches. Specifically, the Basalt~\cite{Basalt_Usenko2020}
implementation performs tracking on the gray-scale intensity
$I_t$ and optimizes an $\mathrm{SE}(2)$ transform $\mathbf{T}_{\text{2D}}$ per patch
$\mathcal{P} \equiv \mathcal{N}(\mathbf{p})$ centered on a keypoint $\mathbf{p}$

\begin{equation}
    \mathbf{T}_{\text{2D}} = \begin{pmatrix}
    \mathbf{R}_\text{2D} & \mathbf{t}_\text{2D} \\
    \mathbf{0} & 1 \end{pmatrix},
\end{equation}

by minimizing the photometric error
\begin{equation}
    \mathbf{R}_\text{2D}, \; \mathbf{t}_\text{2D}
    = \arg \min_{\mathbf{R}_\text{2D}, \;
    \mathbf{t}_\text{2D}}
    \sum_{\tilde{\mathbf{p}} \in \mathcal{P}}
    \left(\frac{I_{t-1}(\tilde{\mathbf{p}})}{\bar{I}_{t-1}}
    - \frac{I_{t}(\mathbf{R}_\text{2D} \tilde{\mathbf{p}}
    + \mathbf{t}_\text{2D})}{\bar{I}_{t}} \right)^2,
\end{equation}
where $\bar{I}_{t-1}$ and $\bar{I}_{t}$ are mean patch
intensities that normalize for illumination change. The
translation $\mathbf{t}_\text{2D}$ corresponds to
an estimate of the optical flow $\mathbf{v}$, while the rotational component
$\mathbf{R}_\text{2D}$ captures local image rotation that a
purely translational model, such as a codec motion vector, cannot represent.

Directly replacing KLT-Tracking with decoded motion vectors is
not a viable alternative: as discussed above, motion vectors are
optimized for compression rather than motion fidelity, making
them noisy estimates of true scene motion. Moreover, they are
purely translational and cannot represent the in-plane rotation
captured by the $\mathbf{R}_\text{2D}$ component of the
$\mathrm{SE}(2)$ model.

However, motion vectors can bridge
displacements far larger than what image pyramids alone can
recover, making them well suited for use as initialization priors.
Since $\mathbf{t}_\text{2D}$ points forward in time (matching
the convention of $\mathbf{v}$) and
$\mathbf{v} \approx -\mathbf{d}_k$, we initialize
$\mathbf{t}_\text{2D}$ with the negated motion vector
$-\mathbf{d}_k$, while retaining the identity for
$\mathbf{R}_\text{2D}$, and let the KLT optimizer refine both
from this starting point. For backward
tracking from frame $t$ to $t-1$, used for cycle-consistency
verification, the initialization is $\mathbf{d}_k$ directly.

As \cref{fig:guess_error} shows, for successful tracks, this initialization reduces the distances from the initialization points to the final converged positions when using motion vectors; the majority of motion vectors lie within $1$--$2$
pixels of the converged solution, confirming their effectiveness
as initialization priors. Note that video compression
additionally introduces artifacts into the decoded images (see Fig.~\ref{fig:artifacts}),
adding further noise to the optimization, thereby reinforcing the
need for a full KLT refinement rather than relying on motion
vectors directly.

\subsection{Bridging I-Frames}
\label{subsec:i-frames}
As established in \cref{sec:video_coding}, I-frames carry no
inter-frame prediction and therefore yield no motion vectors.
This is particularly challenging because I-frames tend to occur
precisely when scene content changes substantially, the same
conditions under which establishing feature correspondences is
most difficult. To maintain motion-vector initialization across
I-frame boundaries, we reuse the motion vectors
$\mathbf{d}_k$ decoded from the last available P-frame, assuming constant motion. While
these vectors do not reflect the true displacement at the
I-frame, they provide a temporally proximate estimate that,
in many empirical cases, lies sufficiently close to the true optical flow to
place the KLT optimizer within its convergence basin.

\subsection{Outliers through False Positives and Dynamic Objects}
\label{subsec:outliers}

As the motion-vector initialization can cover large displacements, the tracking is not constrained to the same locality as the previous-pixel prior baseline. This has the potential to introduce two kinds of false positive tracks:
\begin{itemize}
    \item A wrong motion vector can point in the wrong direction, but represent visually similar image regions (e.g., wall area of \cref{fig:codec})
    \item A correct motion vector is associated with a dynamic object (see \cref{fig:codec} on the arm). These features are difficult to track because of their large displacement.
\end{itemize}
If, in both of these scenarios, the forward tracking is already close to the minima, then the backward track will be initialized close to its origin. As the motion vectors from the video encoder represent photometrically similar regions, the traditional forward-backward consistency of KLT-Tracking is not reliable in filtering out these outliers. These correspondences badly influence the bundle adjustment in the backend and cause drift or divergence in the trajectory.

To reduce their influence, we propose the following strategy: track each point \textit{with} and \textit{without} motion priors. If a point is tracked only once, we treat it as a success. If both approaches successfully track the point, we check whether they are consistent and reject any that are inconsistent. This allows us to identify possible outliers in the tracks.

\subsection{Implementation Details}
As discussed in \cref{sec:intro}, KLT-Tracking has strong assumptions and is not well suited for compressed data. In consequence, this also holds for our foundation, Basalt \cite{Basalt_Usenko2020}, with updates from \cite{de2025monado}. In addition to motion-vector initialization, we adapt the original system in two ways: First, we relax the threshold for the forward-backward consistency, allowing for more tracks over the whole sequence. Second, we adapt the patch $\mathcal{P}$ to cover a larger area. This captures more of the image structure, increasing robustness. The reported numbers for Basalt include these changes.
\section{Experiments}
\label{sec:experiments}

\begin{table}[t]\fontsize{7}{7}\selectfont
    \centering
    \caption{\textbf{EuRoC dataset.} Following \cite{de2025monado}, we use thresholds for ATE and RTE ($\Delta=6$ frames) of $10\,\mathrm{m}$ and $10\,\mathrm{cm}$, respectively, and mark runs that exceed these thresholds as divergent with $\infty$. Incomplete runs (crashes) are marked as $\times$. Systems that are unable to estimate more than $50\%$ of the trajectory without resetting are marked as R. We show median (MED) and average (AVG). For avoiding outliers skewing the AVG, we exclude sequences that fail in any of the systems with reasonable metrics. We mark these sequences in \textbf{\color{gray}gray}. MoV-SLAM does not produce reasonable ATE and RTE, and ORB-SLAM3 does not produce reasonable RTE.
    Only on easy sequences does Basalt show slightly better performance than VOCA. On EuRoC, our method improves the mean RTE by $\approx 15\%$ and the mean ATE by $\approx 10\%$, while being more robust, as shown by the number of successfully completed sequences.}
    \begin{tabular}{r R{26pt}R{26pt}R{26pt}R{26pt}R{26pt} p{6pt} R{26pt}R{26pt}R{26pt}R{26pt}R{26pt}}
        \toprule
        & \multicolumn{5}{c}{ATE [cm] (SE3 aligned)} &  & \multicolumn{5}{c}{RTE [cm] ($\mathrm{\Delta} = 6$ frames)}  \\
        \cmidrule{2-6} \cmidrule{8-12}
         & \multicolumn{1}{c}{MoV-} & \multicolumn{1}{c}{ORB-} & & & \multicolumn{1}{c}{VOCA} & & \multicolumn{1}{c}{MoV-} & \multicolumn{1}{c}{ORB-} & & & \multicolumn{1}{c}{VOCA} \\
        & \multicolumn{1}{c}{SLAM}  & \multicolumn{1}{c}{SLAM3} & \multicolumn{1}{c}{OKVIS2} & \multicolumn{1}{c}{Basalt} & \multicolumn{1}{c}{\textbf{OURS}} & & \multicolumn{1}{c}{SLAM}  & \multicolumn{1}{c}{SLAM3} & \multicolumn{1}{c}{OKVIS2} & \multicolumn{1}{c}{Basalt} & \multicolumn{1}{c}{\textbf{OURS}}\\
\midrule
EMH01 & $\infty$ & \textbf{3.80} & \underline{12.10} & 16.20 & 17.50 & & $\infty$ & $\infty$ & 1.371 & \textbf{0.595} & \underline{0.637} \\
EMH02 & $\infty$ & \textbf{5.30} & 13.60 & \underline{10.60} & 13.30 & & $\infty$ & $\infty$ & 1.528 & \underline{0.621} & \textbf{0.604} \\
EMH03 & 373.90 & \textbf{4.70} & \underline{17.60} & 20.00 & 20.60 & & $\infty$ & $\infty$ & 2.950 & \underline{1.504} & \textbf{1.487} \\
EMH04 & \color{gray}698.50 & \color{gray}949.30 & \color{gray}$\infty$ & \color{gray}\textbf{21.90} & \color{gray}\underline{22.90} & & \color{gray}$\infty$ & \color{gray}$\infty$ & \color{gray}$\infty$ & \color{gray}\underline{2.095} & \color{gray}\textbf{2.075} \\
EMH05 & $\times$ & \textbf{12.30} & 20.50 & 24.30 & \underline{20.10} & & $\times$ & $\infty$ & 4.041 & \textbf{1.603} & \underline{1.656} \\
EV101 & 681.90 & \textbf{9.40} & 15.60 & \underline{10.10} & 11.30 & & $\infty$ & $\infty$ & 1.899 & \underline{1.423} & \textbf{1.405} \\
EV102 & \color{gray}484.40 & \color{gray}R & \color{gray}\underline{20.60} & \color{gray}$\times$ & \color{gray}\textbf{8.30} & & $\color{gray}\infty$ & \color{gray}R & \color{gray}\underline{3.176} & \color{gray}$\times$ & \color{gray}\textbf{1.549} \\
EV103 & 245.00 & 138.30 & 66.80 & \underline{51.40} & \textbf{38.30} & & $\infty$ & $\infty$ & 8.228 & \underline{6.185} & \textbf{4.085} \\
EV201 & 781.50 & 42.40 & 11.00 & \textbf{7.60} & \underline{8.10} & & $\infty$ & $\infty$ & \textbf{1.582} & \underline{1.981} & 2.031 \\
EV202 & 549.50 & \underline{12.50} & 20.20 & 13.70 & \textbf{9.60} & & $\infty$ & $\infty$ & 3.073 & \underline{1.667} & \textbf{1.446} \\
EV203 & \color{gray}203.30 &\color{gray}R & \color{gray}533.20 & \color{gray}$\infty$ & \color{gray}\textbf{185.60} & & \color{gray}$\infty$ & \color{gray}R & \color{gray}$\infty$ & \color{gray}$\infty$ & \color{gray}$\infty$ \\
\midrule
    AVG & $\infty$ & 28.59 &  22.18 & \underline{19.24} & \textbf{17.35} & & $\infty$ & $\infty$ & 3.084 & \underline{1.947} & \textbf{1.669} \\
    MED & 681.90 & \textbf{12.50} & 20.20 & 20.00 & \underline{17.50} & & $\infty$ & $\infty$ & 3.073 & \underline{1.667} & \textbf{1.549} \\
         \bottomrule
    \end{tabular}
    \label{tab:qualitative_euroc}
\end{table}

We evaluate on three different \ac{VSLAM} datasets representative of use cases with limited compute due to hardware constraints. EuRoC \cite{burri2016euroc} represents drone data captured in two different scenarios, a machine hall and a standard room. TUM-VI \cite{schubert2018tum} represents handheld motion commonly found in smartphone or body-cam footage. The \ac{MSD} \cite{de2025monado} is a recent dataset captured with a head-mounted sensor setup for applications such as VR and humanoid robotics, and is much longer than the previous two.

We use sequence acronyms with the first letter representing the dataset; e.g., EMH04 for \texttt{MH\_04\_difficult} from EuRoC, TR2 for \texttt{room2} from TUM-VI, and MIPB05 for \texttt{MIPB05\_beatsaber\_fitbeat\_360\_expert} from \ac{MSD}.

We compare against the following baselines: \textbf{1. ORB-SLAM3}~\cite{ORB_SLAM3_campos2021} is a state-of-the-art descriptor-based visual \ac{SLAM} system that is not designed with compressed videos in mind. Similarly, \textbf{2. OKVIS2}~\cite{leutenegger2022okvis2} presents another \ac{SLAM} system with strong performance. \textbf{3. MoV-SLAM}~\cite{turner2023mov} presents the only work in the literature that is specifically designed for compressed videos. \textbf{4. Basalt}~\cite{Basalt_Usenko2020} is the baseline upon which we build VOCA and is therefore conceptually the closest, as it also uses KLT-Tracking correspondences. We evaluate our method in a purely stereo \ac{VO} setting. To simulate real-time usage, we report causal estimates only. We disable loop closure for ORB-SLAM3 and OKVIS2 to isolate the front-end tracking.

\begin{figure}[t!]
  \centering
  \includegraphics[trim={0 0.0cm 0 0},clip,width=0.98\linewidth]{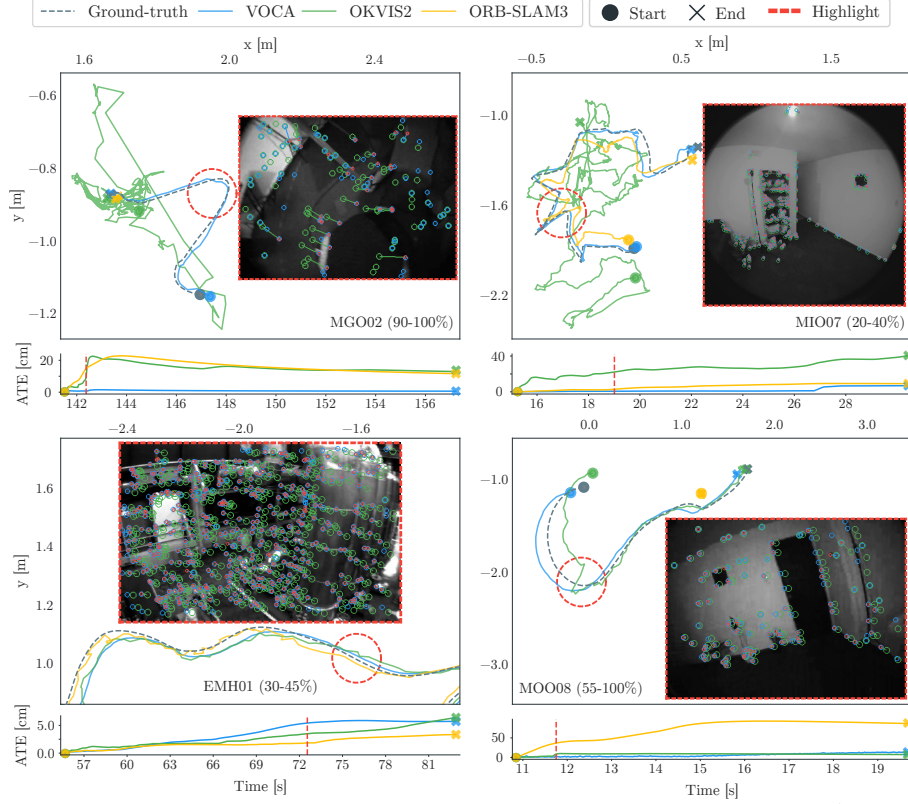}
  \vspace{-1em}
  \caption{\textbf{Qualitative trajectory examples.} To showcase the quality of
  VOCA trajectories, in addition to \cref{fig:trajectories}, we provide segments covering more devices and
  scenarios. We highlight specific frames with dashed red outlines. Inside each frame,
  we draw the \textbf{\mdred optical-flow} solutions, if any, the \textbf{\mdgreen feature
  location} in the previous frame used by standard optical-flow-based
  feature trackers, and the \textbf{\mdblue motion-vector} prior extracted from the H.264
  decoder used by VOCA. As can be seen in the images, our prior is closer to the solution in almost all cases, even for large displacements such as those seen
  in MGO02 (from MSD).}
  \label{fig:trajectories-appendix}
\end{figure}

\begin{table}[t]\fontsize{7}{7}\selectfont
    \centering
    \caption{\textbf{TUM-VI dataset.} We restrict the evaluation to the room sequences, as they are the only ones with full ground truth. For almost all sequences, VOCA clearly outperforms the baseline Basalt~\cite{Basalt_Usenko2020}. The difference is more pronounced on the most challenging sequence, TR5, with an ATE improvement of $\approx 85\%$. Only ORB-SLAM3 achieves comparable performance in global consistency (ATE) but exceeds the RTE threshold across all sequences. Symbols and shading follow the conventions of \cref{tab:qualitative_euroc}.}
\begin{tabular}{r R{26pt}R{26pt}R{26pt}R{26pt} p{6pt} R{26pt}R{26pt}R{26pt}R{26pt}}
    \toprule
    & \multicolumn{4}{c}{ATE [cm] (SE3 aligned)} &  & \multicolumn{4}{c}{RTE [cm] ($\mathrm{\Delta} = 6$ frames)}  \\
    \cmidrule{2-5} \cmidrule{7-10}
    & \multicolumn{1}{c}{ORB-} & & & \multicolumn{1}{c}{VOCA} & & \multicolumn{1}{c}{ORB-} & & & \multicolumn{1}{c}{VOCA} \\
    & \multicolumn{1}{c}{SLAM3} & \multicolumn{1}{c}{OKVIS2} & \multicolumn{1}{c}{Basalt} & \multicolumn{1}{c}{\textbf{OURS}} & & \multicolumn{1}{c}{SLAM3} & \multicolumn{1}{c}{OKVIS2} & \multicolumn{1}{c}{Basalt} & \multicolumn{1}{c}{\textbf{OURS}}\\
    \midrule
TR1 & \color{gray}7.70 & \color{gray}$\times$ & \color{gray}\bfseries{6.30} & \color{gray}\underline{6.80} &  & \color{gray}$\infty$ & \color{gray}$\times$ & \color{gray}\underline{1.010} & \color{gray}\bfseries{0.896} \\
TR2 & \bfseries{7.60} & \underline{12.00} & 12.80 & 12.70 &  & $\infty$ & 1.489 & \bfseries{0.941} & \bfseries{0.941} \\
TR3 & \bfseries{12.30} & 21.30 & \underline{15.30} & 15.90 &  & $\infty$ & 2.028 & \underline{0.910} & \bfseries{0.853} \\
TR4 & \underline{8.10} & 8.70 & 14.40 & \bfseries{7.10} &  & $\infty$ & \underline{2.085} & 4.458 & \bfseries{0.733} \\
TR5 & \underline{7.80} & 54.80 & 54.50 & \bfseries{7.70} &  & $\infty$ & \underline{3.907} & 7.287 & \bfseries{1.046} \\
TR6 & 7.80 & 5.80 & \bfseries{4.70} & \underline{4.80} &  & $\infty$ & 1.091 & \bfseries{0.502} & \underline{0.503} \\
\midrule
AVG & \bfseries{8.72} & 20.52 & 20.34 & \underline{9.64} &  & $\infty$ & \underline{2.120} & 2.820 & \bfseries{0.815} \\
MED & \underline{7.80} & 16.65 & 13.60 & \bfseries{7.40} &  & $\infty$ & 2.057 & \underline{0.975} & \bfseries{0.874} \\
    \bottomrule
\end{tabular}
\label{tab:qualitative_tim_vi}
\end{table}
Whenever possible, we provide the same input data. However, MoV-SLAM and ORB-SLAM3 require stereo-rectified images for certain camera models, introducing computational overhead. As TUM-VI and \ac{MSD} employ fisheye cameras with large distortions, MoV-SLAM is not reported, as it cannot work without rectification and would be heavily disadvantaged.
We run the official MoV-SLAM implementation on EuRoC and report those metrics since, despite our best efforts, we were unable to reproduce the numbers claimed in \cite{turner2023mov}.
We compress each camera separately with a maximum bitrate of 500 kbps when not otherwise specified, using the popular tool \texttt{ffmpeg}. We provide usage examples in the supplementary material. We adapt ORB-SLAM3, OKVIS2, and Basalt to support video input. MoV-SLAM expects a video of alternating left and right frames. This allows it to compute not only temporal but also stereo correspondences. To simulate real-world live applications, we \textit{do not} use bidirectional encoding for video compression, giving us purely causal data. Furthermore, we evaluate the causal pose predictions of the methods, i.e., the poses of each frame after its first optimization, before future frames become available. Qualitative examples of trajectories are shown in \cref{fig:trajectories} and \cref{fig:trajectories-appendix}. We evaluate on the common \textit{absolute trajectory error} (ATE) and \textit{relative trajectory error} (RTE) metrics. Details on the metrics can be found in the supplementary material.

\subsection{EuRoC}
On EuRoC (see \cref{tab:qualitative_euroc}), we show competitive performance against established VO systems. Especially on the hard sequences (EMH04, EMH05, EV103, EV203), we show significant improvements over most baselines. In the RTE, we outperform the closest baseline, Basalt, by $\approx 15\%$, demonstrating superior relative tracking performance. VOCA recovers EV102, whereas most systems fail.

\subsection{TUM-VI}
On the TUM-VI room sequences (\cref{tab:qualitative_tim_vi}), we show consistent improvement over the baselines. We exclude MoV-SLAM due to the previously discussed problems with fisheye cameras. Unlike Basalt, VOCA performs consistently across sequences. The strong prior information of the motion vectors allows VOCA to overcome difficult sections like the ones in TR5, reducing both ATE and RTE.

\subsection{Monado SLAM Dataset}

\begin{table}[t]\fontsize{7}{7}\selectfont
    \centering
    \caption{\textbf{\ac{MSD} dataset.} Given \ac{MSD}'s small room environment, we reduce the divergence threshold for ATE to $1\,\mathrm{m}$. The high reset count of ORB-SLAM3 (see \cref{fig:heatmap}) makes it necessary to exclude it from the AVG aggregate. We report the mean, median, and percentage of successful sequences (success rate, SR) over the whole dataset. Rankings for individual sequences can be found in \cref{fig:heatmap}, and the quantitative results can be found in \cref{fig:trajectories-appendix}. \ac{MSD} is the most challenging dataset, with a combination of difficult movements and dynamic objects. This is the only dataset that exhibits consistent failures across all methods. Compared to the second-best method, VOCA improves median ATE by $37\%$, median RTE by $40\%$, and ATE SR by 13 percentage points ($36\%$ relative improvement).}
    \begin{tabular}{r R{28pt}R{28pt}HR{28pt}R{28pt} p{6pt} R{28pt}R{28pt}HR{28pt}R{28pt}}
        \toprule
        & \multicolumn{5}{c}{ATE [cm] (SE3 aligned)} &  & \multicolumn{5}{c}{RTE [cm] ($\mathrm{\Delta} = 6$ frames)}  \\
        \cmidrule{2-6} \cmidrule{8-12}
        & \multicolumn{1}{c}{ORB-} & & & & \multicolumn{1}{c}{VOCA} & & \multicolumn{1}{c}{ORB-} & & & & \multicolumn{1}{c}{VOCA} \\
        & \multicolumn{1}{c}{SLAM3} & \multicolumn{1}{c}{OKVIS2} & & \multicolumn{1}{c}{Basalt} & \multicolumn{1}{c}{\textbf{OURS}} & & \multicolumn{1}{c}{SLAM3} & \multicolumn{1}{c}{OKVIS2} & & \multicolumn{1}{c}{Basalt} & \multicolumn{1}{c}{\textbf{OURS}}\\
        \midrule
        AVG & R & 38.17 & & \underline{5.51} & \bfseries{4.46} & &  R & 5.223 & & \underline{1.478} & \bfseries{0.755} \\
        MED & R & \underline{124.70} & & 479.20 & \bfseries{78.30} &  & R & \underline{8.692} & & 29.124 & \bfseries{5.212} \\
        SR & \underline{34.4$\%$} & 31.2$\%$ &  & \underline{34.4$\%$} & \bfseries{46.9$\%$} &  & 21.9$\%$ & \underline{48.8$\%$} & & 39.1$\%$ & \bfseries{53.1$\%$} \\
         \bottomrule
    \end{tabular}
    \label{tab:qualitative_monado}
\end{table}
\begin{figure}[t]
  \centering
  \includegraphics[width=\linewidth]{images/msd_heatmap.pdf}
  \caption{\textbf{\ac{MSD} overview.} We indicate divergences and resets with gray $\infty$ and $\circlearrowright$, respectively. First-place ranks for each sequence are shown in green, and non-first ranks are shown in blue. See the supplementary for the detailed numbers of each run.}
  \label{fig:heatmap}
\end{figure}
Similar to TUM-VI, the reduced camera overlap means stereo-rectification would produce extremely narrow fields of view, penalizing performance disproportionately for MoV-SLAM \cite{turner2023mov}, so it is also excluded from this evaluation. Overall, the dataset is significantly harder than EuRoC and TUM-VI due to regular dynamic occlusions, strong accelerations, and fast rotations. In \cref{tab:qualitative_monado}, we report metrics over all sequences; all methods show a significantly worse RTE compared to other datasets. We show an overview of each sequence in \cref{fig:heatmap} with the full numbers reported in the supplementary material. It is also the only dataset where methods either do not complete a sequence or produce results that are beyond the selected thresholds. Due to this greater difficulty, in \cref{tab:qualitative_monado}, our work shows significant improvements when compared to the second-best method in each metric. Over the whole dataset, the median ATE and RTE improve by around $37\%$ and $40\%$. In terms of the number of successfully estimated sequences (success rate) for non-diverging ATE results, VOCA shows an improvement of $36\%$.

\subsection{Performance vs. Compression}
\label{subsec:performance_vs_compression}

\begin{figure}[tb]
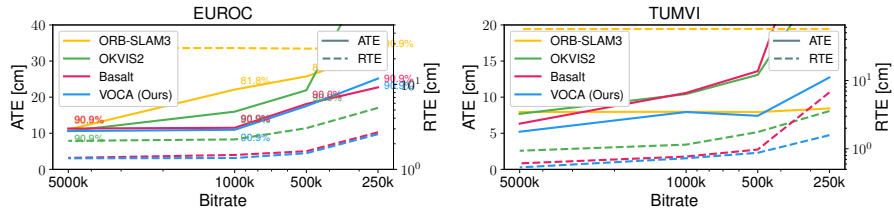

  \centering
  \begin{subfigure}[b]{0.48\textwidth}
    \includegraphics[width=\linewidth]{images/euroc.pdf}
  \end{subfigure}
  \begin{subfigure}[b]{0.48\textwidth}
    \includegraphics[width=\linewidth]{images/tumvi.pdf}
  \end{subfigure}
  \caption{
    \textbf{Median ATE/RTE vs. Bitrate.} Lower bitrates lead to smaller file sizes. Numbers indicate a success rate of less than $100\%$ for ATE. VOCA shows stable performance even at low bitrates. As a tracking-based system, Basalt \cite{Basalt_Usenko2020} suffers from compression artifacts, especially on the TUM-VI dataset. ORB-SLAM3 \cite{ORB_SLAM3_campos2021} shows steady performance in ATE on TUM-VI but delivers poor performance in RTE. This indicates good global consistency but poor frame-to-frame tracking.
  }
  \label{fig:performance_vs_compression}
\end{figure}
In \cref{fig:performance_vs_compression}, we investigate how different compression bitrates affect the tracking systems on EuRoC and TUM-VI. OKVIS2 suffers especially heavily from compression. Overall, VOCA is the most stable VO system across both datasets. VOCA's performance is relatively stable up to 500 kbps, which represents about $\times 100$ compression for most datasets. RTE, in particular, shows quite stable results.

\subsection{Ablation Study}
\label{subsec:ablation}
\begin{table}[t]\fontsize{7}{7}\selectfont
    \centering
    \caption{\textbf{Ablation study.} We evaluate three motion-vector (MV) integration strategies in combination with optical flow (OF) on EuRoC and TUM-VI: (A)~OF $\Rightarrow$ MV tracks without priors first and falls back to MV priors; (B) reverses this order; (C)~MV $||$ OF runs both in parallel, keeping single-mode tracks directly and filtering dual-mode tracks by consistency. +~IF denotes I-frame bridging. (C)\,+\,IF is our method, VOCA.}
    \begin{tabular}{rl R{22pt}R{22pt}HR{22pt}R{22pt}R{22pt} p{6pt} R{22pt}R{22pt}HR{22pt}R{22pt}R{22pt}}
        \toprule
        & & \multicolumn{6}{c}{ATE [cm] (SE3 aligned)} &  & \multicolumn{6}{c}{RTE [cm] ($\mathrm{\Delta} = 6$ frames)}  \\
        \cmidrule{3-8} \cmidrule{10-15}
        & & \multicolumn{1}{r}{} & \multicolumn{1}{r}{} & \multicolumn{1}{r}{} & \multicolumn{1}{r}{} & \multicolumn{1}{r}{} & \multicolumn{1}{r}{(C)} & & \multicolumn{1}{r}{} & \multicolumn{1}{r}{} & \multicolumn{1}{r}{} & \multicolumn{1}{r}{} & \multicolumn{1}{r}{} & \multicolumn{1}{r}{(C)} \\
        & & \multicolumn{1}{r}{Basalt} & \multicolumn{1}{r}{(A)} & \multicolumn{1}{r}{} & \multicolumn{1}{r}{(B)} & \multicolumn{1}{r}{(C)} & \multicolumn{1}{r}{+ IF} & & \multicolumn{1}{r}{Basalt} & \multicolumn{1}{r}{(A)} & \multicolumn{1}{r}{} & \multicolumn{1}{r}{(B)} & \multicolumn{1}{r}{(C)} & \multicolumn{1}{r}{+ IF} \\
        \midrule
EuRoC
& AVG & 19.50 & \bfseries{17.20} & \underline{17.40} & 18.40 & 18.70 & \underline{18.00} &  & 1.964 & 1.871 & 1.911 & 2.044 & \bfseries{1.710} & \underline{1.714} \\
& MED & 20.00 & \bfseries{12.80} & \underline{12.80} & \underline{16.80} & 17.30 & 17.50 &  & 1.667 & 1.608 & 1.608 & 1.656 & \underline{1.552} & \bfseries{1.549} \\
        \midrule
TUM-VI
& AVG & 18.00 & 10.40 & \underline{10.10} & 10.70 & \bfseries{9.10} & \underline{9.20} &  & 2.518 & 1.380 & 1.221 & 1.121 & \underline{0.919} & \bfseries{0.829} \\
& MED & 13.60 & 10.30 & 10.10 & 10.60 & \underline{8.90} & \bfseries{7.40} &  & 0.975 & 1.234 & 0.968 & 0.968 & \bfseries{0.872} & \underline{0.874} \\
        \bottomrule
    \end{tabular}%
    \label{tab:ablation}
    \vspace{-1em}
\end{table}

We consider three strategies for integrating \ac{MV} into tracking,
balancing the strength of the MV prior against the risk of propagating outliers:
\begin{enumerate}[label=(\Alph*)]
    \item \textbf{OF $\Rightarrow$ MV} tracks first with the base same-pixel prior, falling back to
    MV priors for failed points, assuming most points are easy to track and MV priors are
    only needed in difficult sections.
    \item \textbf{MV $\Rightarrow$ OF} reverses this order, placing high confidence
    in the MV prior and resorting to prior-free tracking only when it fails.
    \item \textbf{MV $||$ OF} is our default strategy from \cref{subsec:outliers}. Both modes
    run in parallel; tracks successful in a single mode are kept; tracks successful in both modes are kept only if mutually
    consistent.
\end{enumerate}
For (C), we additionally evaluate bridging I-frame keyframes (+~IF), as described in
\cref{subsec:i-frames}. \cref{tab:ablation} shows that (C)~MV $||$ OF achieves the best overall
performance, with (A)~OF $\Rightarrow$ MV close behind. Intuitively, most data represents easy motion where points are straightforward to track. MV priors become essential only in difficult
sections. Adding I-frame bridging further improves (C), yielding the best median ATE on
TUM-VI and competitive RTE across both datasets.
\section{Conclusion}
\label{sec:conclusion}

We present VOCA, a \acl{VO} system with codec awareness. By leveraging motion vectors
and I-frame structure from the video codec to aid KLT-Tracking, VOCA overcomes difficult sections, achieving more accurate and stable trajectories. This is particularly apparent in RTE, a metric critical for mixed reality, where local smoothness directly impacts user experience.
Codec-aware tracking is equally relevant for autonomous systems where bandwidth and memory constraints
make compression unavoidable, such as drones, wearables, and robotic assistants under a compress-then-analyze design
\cite{hoferH264CompressThenAnalyzeTransmission2023}. Unlike conventional SLAM
systems, VOCA remains robust under
heavy compression. With the vast majority of video data on the internet being
codec-compressed, codec-aware methods like VOCA are essential for efficiently unlocking
large amounts of data for downstream vision and learning tasks. We hope this work inspires
further research on leveraging information from codecs in computer vision tasks.

\textbf{Acknowledgments.} This work was supported by the European Research
Council (ERC) Advanced Grant SIMULACRON, by the DFG project CR 250/26-1
“4D-YouTube”, by the GNI Project “AI4Twinning”, and by the Munich Center for
Machine Learning.

\bibliographystyle{splncs04}
\bibliography{main}

\clearpage
\setcounter{tocdepth}{2}
\setcounter{page}{1}
\appendix

\section{Metrics}

In this paper, we utilize two commonly used metrics to evaluate the performance
of Visual Odometry algorithms: the Absolute Trajectory Error (ATE) and the
Relative Trajectory Error (RTE). We provide a review of them and discuss their
advantages and downsides. In what follows, we denote the estimated pose at a
timestep $t \in [1..N]$ as
\begin{equation}
    \estimated_t =
    \begin{pmatrix}
        \estimatedR_t & \estimatedT_t \\
        \mathbf{0}    & 1             \\
    \end{pmatrix}
\end{equation}
and the ground truth pose at the same timestep as
\begin{equation}
    \groundtruth_t =
    \begin{pmatrix}
        \groundtruthR_t & \groundtruthT_t \\
        \mathbf{0}      & 1               \\
    \end{pmatrix} \,.
\end{equation}
The translational error between two poses is given by the Euclidean norm of the translational difference:
\begin{equation}
    e_\mathrm{t}(\mathbf{T}_i, \mathbf{T}_j) =  \|\mathbf{t}_j - \mathbf{t}_i\|_2 \,,
\end{equation}

For all methods, we evaluated the metrics considering every frame instead of
method-dependent keyframes. Only for methods such as ORB-SLAM3 that are able to
create multiple maps, we restrict the evaluation to the timesteps of the largest
maps. For ORB-SLAM3, new maps are created when tracking of the last map has
failed. New maps are created with their initial pose (re-)set at the origin.

\subsection{Absolute trajectory error (ATE)}

The ATE metric requires first aligning the estimated trajectory to the ground
truth. Since we are evaluating \textit{stereo} Visual Odometry algorithms, we do
not perform scale alignment. We find the optimal rigid body transformation
$\mathbf{T}_{\mathrm{align}}$ that minimizes the quadratic error between
the estimated and ground-truth poses. It is commonly computed with the Umeyama
algorithm. The ATE is now given by the root mean square error (RMSE) of the
translational error over each pose estimate:
\begin{equation}
    \mathrm{ATE} = \sqrt{\frac{1}{N} \sum_{i=1}^{N} \| e_\mathrm{t}(\groundtruth_i, \mathbf{T}_{\mathrm{align}} \estimated_i) \|^2}
\end{equation}

The ATE is a good metric to capture global consistency of a trajectory and its
long-term drift. However, the alignment process and the quadratic nature of the
metric result in high sensitivity to sparse outliers and rotation estimation.
As the metric can be dominated by a few timesteps, it is not sufficient to
evaluate local estimation quality.

\subsection{Relative trajectory error (RTE)}

The relative trajectory error does not need alignment between the trajectories.
Instead, it measures local drift between the estimates on a per-segment basis. It is
defined as the RMSE of the relative poses between two timesteps that are
$\Delta$ frames apart:
\begin{equation}
    \mathrm{RTE} = \sqrt{\frac{1}{M} \sum_{i=1}^{M - 1} \| e_\mathrm{t}(\groundtruth_{i\Delta}^{-1}\groundtruth_{(i + 1)\Delta}, \estimated_{i\Delta}^{-1}\estimated_{(i + 1)\Delta}) \|^2} \,,
\end{equation}

with $M = \lfloor N / \Delta \rfloor$.

We follow the literature by choosing $\Delta = 6$ frames. As the RTE looks at close
frames, it is useful for representing local consistency. As can be seen in the
main paper, especially ORB-SLAM3 has a high RTE over all datasets. This is
reflected in the qualitative results shown in \cref{fig:trajectories-appendix},
where its trajectory exhibits significant jitter. A downside of the RTE is that
it is not very strong at capturing long-term drift. To address this, one can
average over all possible $\Delta$, but this introduces quadratic complexity
with regard to the trajectory length $N$ and is thus not common in practice.

\section{Video Encoding}

For each camera in each sequence, we use the same encoding strategy, leveraging
the popular and open-source \texttt{ffmpeg} tool. To achieve a constant bitrate,
we perform a two-pass encoding. The first pass gathers image statistics that are
then used in the second pass for doing the actual encoding. The command for the
first pass is given by:
\begin{verbatim}
ffmpeg -y -f concat -safe 0 -r <FRAMERATE> -i <IMG_DIR> \
    -c:v libx264 -b:v <BITRATE> -pix_fmt yuv420p -r <FRAMERATE> \
    -vf setpts=PTS-STARTPTS -x264opts \
    partitions=p8x8,p4x4,i8x8:keyint=1000:\
    me=umh:merange=64:subme=6:bframes=0:ref=1 \
    -passlogfile <PASSLOGFILE> -pass 1 \
    -an -f null /dev/null
\end{verbatim}
and then we perform the second pass with:
\begin{verbatim}
ffmpeg -y -f concat -safe 0 -r <FRAMERATE> -i <IMG_DIR> \
    -c:v libx264 -b:v <BITRATE> -pix_fmt yuv420p -r <FRAMERATE> \
    -vf setpts=PTS-STARTPTS -x264opts \
    partitions=p8x8,p4x4,i8x8:keyint=1000:\
    me=umh:merange=64:subme=6:bframes=0:ref=1 \
    -passlogfile <PASSLOGFILE> -pass 2 \
    -an -f mp4 <OUTPUT>.mp4
\end{verbatim}

We set \verb|<BITRATE>| to \verb|500k| for our experiments to achieve a
compression ratio ranging from around 35$\times$ in MSD to approximately
70$\times$ in EuRoC and TUM-VI. For the experiments in
\cref{subsec:performance_vs_compression}, we replace \verb|<BITRATE>| with
\verb|250k|, \verb|1000k|, and \verb|5000k|, respectively.

\subsection{MoV-SLAM}

For MoV-SLAM, we follow their \verb|README| to encode the stereo video via
\begin{verbatim}
ffmpeg -framerate <FRAMERATE> -pattern_type glob \
    -i <IMG_DIR/*.png> -c:v libx264 -preset "fast" -tune "film" \
    -x264opts "partitions=p8x8,p4x4,i8x8:keyint=1000:\
    me=umh:merange=64:subme=6:bframes=0:ref=1" -f <OUTPUT>.mp4
\end{verbatim}
where the images of the left and right frames are stored in an alternating manner in
the \verb|<IMG_DIR>| as \verb|*_left.png| and \verb|*_right.png|. In contrast to
our approach, this encodes both cameras in a single image stream.

\section{Evaluation on the Monado SLAM Dataset}

\providecommand{\graytext}[1]{\textcolor{gray}{#1}}

\begin{table}[t]\fontsize{7}{7}\selectfont
  \centering
  \caption{\textbf{\ac{MSD} - Valve Index MIO sequences.} VOCA shows
    state-of-the-art results on \ac{MSD}. We follow the same conventions as for
    \cref{tab:qualitative_euroc}, but we consider ATE divergence at $1\,\mathrm{m}$ due to the smaller environment and do not consider ORB-SLAM3 for the \textbf{\color{gray}exclusion} of sequences from the AVG. We report the success rate (SR) for each method as the proportion of non-failed (divergent, reset, crashed) sequences. While VOCA is
    not winning in every sequence, it shows better overall performance on the
    majority of sequences, especially on the RTE metric.}
  \begin{tabular}{r R{48pt}R{32pt}R{32pt}R{32pt} p{6pt} R{48pt}R{32pt}R{32pt}R{32pt}}
    \toprule
          & \multicolumn{4}{c}{ATE [cm] (SE3 aligned)} &          & \multicolumn{4}{c}{RTE [cm] ($\mathrm{\Delta} = 6$ frames)}                                                                                   \\
    \cmidrule{2-5} \cmidrule{7-10}
          & ORB-SLAM3                                  & OKVIS2   & Basalt                                                       & VOCA            &  & ORB-SLAM3     & OKVIS2        & Basalt   & VOCA            \\
    \midrule
    MIO01 & \graytext{R}                               & \graytext{$\infty$} & \graytext{$\infty$}                                  & \graytext{$\infty$}        &  & \graytext{R}             & \graytext{$\infty$}      & \graytext{$\infty$} & \graytext{$\infty$}        \\
    MIO02 & \graytext{R}                               & \graytext{$\infty$} & \graytext{$\infty$}                                  & \graytext{$\infty$}        &  & \graytext{R}             & \graytext{$\infty$}      & \graytext{$\infty$} & \graytext{$\infty$}        \\
    MIO03 & \graytext{\textbf{51.29}}                  & \graytext{$\infty$} & \graytext{$\infty$}                                  & \graytext{$\infty$}        &  & \graytext{\textbf{8.78}} & \graytext{$\infty$}      & \graytext{$\infty$} & \graytext{$\infty$}        \\
    MIO04 & \graytext{R}                               & \graytext{$\infty$} & \graytext{$\infty$}                                  & \graytext{$\infty$}        &  & \graytext{R}             & \graytext{$\infty$}      & \graytext{$\infty$} & \graytext{$\infty$}        \\
    MIO05 & \graytext{R}                               & \graytext{$\infty$} & \graytext{20.99}                                     & \graytext{\textbf{15.18}}  &  & \graytext{R}             & \graytext{$\infty$}      & \graytext{1.58}     & \graytext{\textbf{0.96}}   \\
    MIO06 & \graytext{R}                               & \graytext{$\infty$} & \graytext{$\infty$}                                  & \graytext{$\infty$}        &  & \graytext{R}             & \graytext{$\infty$}      & \graytext{$\infty$} & \graytext{$\infty$}        \\
    MIO07 & 57.07                                      & 72.29    & 15.79                                                & \textbf{11.17}  &  & 7.29          & 8.44          & 2.19     & \textbf{1.92}   \\
    MIO08 & \graytext{R}                               & \graytext{86.49}    & \graytext{$\infty$}                                  & \graytext{\textbf{65.00}}  &  & \graytext{R}             & \graytext{$\infty$}      & \graytext{$\infty$} & \graytext{\textbf{5.13}}   \\
    MIO09 & 1.13                                       & 35.62    & 0.34                                                 & \textbf{0.26}   &  & 9.07          & 9.84          & 0.21     & \textbf{0.18}   \\
    MIO10 & \graytext{8.71}                            & \graytext{$\infty$} & \graytext{5.18}                                      & \graytext{\textbf{2.84}}   &  & \graytext{$\infty$}      & \graytext{$\infty$}      & \graytext{1.05}     & \graytext{\textbf{0.90}}   \\
    MIO11 & 11.34                                      & 77.51    & 6.08                                                 & \textbf{4.55}   &  & \graytext{$\infty$}      & \graytext{$\infty$}      & \graytext{1.77}     & \graytext{\textbf{1.20}}   \\
    MIO12 & \graytext{R}                               & \graytext{$\infty$} & \graytext{$\infty$}                                  & \graytext{$\infty$}        &  & \graytext{R}             & \graytext{$\infty$}      & \graytext{$\infty$} & \graytext{$\infty$}        \\
    MIO13 & \graytext{R}                               & \graytext{$\infty$} & \graytext{$\infty$}                                  & \graytext{$\infty$}        &  & \graytext{R}             & \graytext{\textbf{6.03}} & \graytext{$\infty$} & \graytext{7.00}            \\
    MIO14 & \graytext{10.97}                           & \graytext{$\infty$} & \graytext{9.04}                                      & \graytext{\textbf{6.34}}   &  & 2.60          & 3.85          & 0.74     & \textbf{0.72}   \\
    MIO15 & \graytext{$\infty$}                        & \graytext{$\infty$} & \graytext{$\infty$}                                  & \graytext{\textbf{78.35}}  &  & \graytext{$\infty$}      & \graytext{\textbf{5.92}} & \graytext{$\infty$} & \graytext{7.15}            \\
    MIO16 & \graytext{R}                               & \graytext{$\infty$} & \graytext{$\infty$}                                  & \graytext{$\infty$}        &  & \graytext{R}             & \graytext{$\infty$}      & \graytext{$\infty$} & \graytext{$\infty$}        \\
    \midrule
    AVG   & R                                          & 61.81    & 7.41                                                 & \textbf{5.33}   &  & R             & 7.38          & 1.04     & \textbf{0.94}   \\
    MED   & R                                          & 145.93   & 2008.64                                              & \textbf{106.25} &  & R             & 13.78         & 126.25   & \textbf{7.08}   \\
    SR    & 37.5\%                                     & 25.0\%   & 37.5\%                                               & \textbf{50.0\%} &  & 25.0\%        & 31.2\%        & 37.5\%   & \textbf{56.2\%} \\
    \bottomrule
  \end{tabular}
  \label{tab:msdmio}
\end{table}

\begin{table}[t]\fontsize{7}{7}\selectfont
  \centering
  \caption{\textbf{\ac{MSD} - Valve Index "Playing" sequences.} This
    set of sequences from MSD provides a realistic VR use-case in which the
    operator plays high-intensity games containing large displacements between
    frames. Despite ORB-SLAM3 showing good ATE values, VR experience tends to
    better correlate with low RTE, a metric in which VOCA provides clear wins. Symbols and shading follow the conventions of \cref{tab:msdmio}. }
  \begin{tabular}{r R{48pt}R{32pt}R{32pt}R{32pt} p{6pt} R{48pt}R{32pt}R{32pt}R{32pt}}
    \toprule
           & \multicolumn{4}{c}{ATE [cm] (SE3 aligned)} &          & \multicolumn{4}{c}{RTE [cm] ($\mathrm{\Delta} = 6$ frames)}                                                                                       \\
    \cmidrule{2-5} \cmidrule{7-10}
           & ORB-SLAM3                                  & OKVIS2   & Basalt                                                       & VOCA           &  & ORB-SLAM3     & OKVIS2          & Basalt        & VOCA          \\
    \midrule
    MIPB01 & \graytext{24.97}                           & \graytext{$\infty$} & \graytext{\textbf{18.52}}                          & \graytext{39.13}          &  & 7.36          & 7.29            & 2.47          & \textbf{1.47} \\
    MIPB02 & \graytext{\textbf{11.39}}                  & \graytext{$\infty$} & \graytext{$\infty$}                                  & \graytext{19.53}          &  & 5.42          & 7.33            & 4.01          & \textbf{1.14} \\
    MIPB03 & \graytext{R}                               & \graytext{$\infty$} & \graytext{$\infty$}                                  & \graytext{$\infty$}       &  & \graytext{R}             & \graytext{6.86}            & \graytext{$\infty$}      & \graytext{\textbf{6.70}} \\
    MIPB04 & \graytext{6.61}                            & \graytext{$\infty$} & \graytext{$\infty$}                                  & \graytext{\textbf{5.57}}  &  & $\infty$      & 6.45            & 5.43          & \textbf{0.60} \\
    MIPB05 & \graytext{R}                               & \graytext{$\infty$} & \graytext{7.01}                                      & \graytext{\textbf{5.74}}  &  & R             & 7.41            & 1.75          & \textbf{0.47} \\
    MIPB06 & \graytext{\textbf{8.36}}                   & \graytext{73.60}    & \graytext{82.80}                                     & \graytext{$\infty$}       &  & \graytext{$\infty$}      & \graytext{6.82}            & \graytext{\textbf{6.67}} & \graytext{$\infty$}      \\
    MIPB07 & R                                          & 44.80    & 11.53                                                & \textbf{5.25}  &  & R             & 7.74            & 0.79          & \textbf{0.77} \\
    MIPB08 & \graytext{R}                               & \graytext{R}        & \graytext{$\infty$}                                  & \graytext{$\infty$}       &  & \graytext{R}             & \graytext{R}               & \graytext{$\infty$}      & \graytext{$\infty$}      \\
    MIPP01 & \graytext{R}                               & \graytext{$\infty$} & \graytext{$\infty$}                                  & \graytext{$\infty$}       &  & \graytext{R}             & \graytext{\textbf{9.03}}   & \graytext{$\infty$}      & \graytext{$\infty$}      \\
    MIPP02 & \graytext{\textbf{36.50}}                  & \graytext{$\infty$} & \graytext{$\infty$}                                  & \graytext{79.59}          &  & \graytext{8.25}          & \graytext{7.88}            & \graytext{$\infty$}      & \graytext{\textbf{6.60}} \\
    MIPP03 & \graytext{R}                               & \graytext{$\infty$} & \graytext{$\infty$}                                  & \graytext{$\infty$}       &  & \graytext{R}             & \graytext{\textbf{8.94}}   & \graytext{$\infty$}      & \graytext{$\infty$}      \\
    MIPP04 & \graytext{$\infty$}                        & \graytext{R}        & \graytext{$\infty$}                                  & \graytext{$\infty$}       &  & \graytext{$\infty$}      & \graytext{R}               & \graytext{$\infty$}      & \graytext{$\infty$}      \\
    MIPP05 & \graytext{\textbf{76.69}}                  & \graytext{$\infty$} & \graytext{$\infty$}                                  & \graytext{$\infty$}       &  & \graytext{8.57}          & \graytext{7.86}            & \graytext{$\infty$}      & \graytext{\textbf{5.30}} \\
    MIPP06 & \graytext{\textbf{72.53}}                  & \graytext{$\infty$} & \graytext{$\infty$}                                  & \graytext{$\infty$}       &  & \graytext{\textbf{8.94}} & \graytext{9.57}            & \graytext{$\infty$}      & \graytext{$\infty$}      \\
    MIPT01 & \graytext{\textbf{9.70}}                   & \graytext{$\infty$} & \graytext{$\infty$}                                  & \graytext{27.52}          &  & 4.32          & 4.87            & 4.41          & \textbf{1.71} \\
    MIPT02 & \graytext{59.08}                           & \graytext{$\infty$} & \graytext{$\infty$}                                  & \graytext{\textbf{20.11}} &  & \graytext{6.14}          & \graytext{4.37}            & \graytext{$\infty$}      & \graytext{\textbf{0.93}} \\
    MIPT03 & \graytext{\textbf{12.58}}                  & \graytext{$\infty$} & \graytext{$\infty$}                                  & \graytext{47.28}          &  & \graytext{5.81}          & \graytext{5.94}            & \graytext{$\infty$}      & \graytext{\textbf{1.70}} \\
    \midrule
    \midrule
    AVG    & R                                          & R        & 11.53                                                & \textbf{5.25}  &  & R             & R               & 3.14          & \textbf{1.03} \\
    MED    & R                                          & R        & 890.91                                               & \textbf{79.59} &  & R             & R               & 32.36         & \textbf{3.50} \\
    SR     & \textbf{58.8\%}                            & 11.8\%   & 23.5\%                                               & 52.9\%         &  & 47.1\%        & \textbf{88.2\%} & 41.2\%        & 64.7\%        \\
    \bottomrule
    \bottomrule
  \end{tabular}
  \label{tab:msdmip}
\end{table}

\begin{table}[t]\fontsize{7}{7}\selectfont
  \centering
  \caption{\textbf{\ac{MSD} - Odyssey+ MOO sequences.} The Odyssey+
    provides the lowest quality set of sensors with highly noisy VGA cameras. VOCA
    remains competitive with other state-of-the-art systems, although its
    improvements are not as clear as for the other devices from this dataset. We
    hypothesize this is due to the aggressive effects of H.264 compression on
    these grainy images, as can be seen in the MOO08 example in \cref{fig:trajectories-appendix}. Symbols and shading follow the conventions of \cref{tab:msdmio}. }
  \begin{tabular}{r R{48pt}R{32pt}R{32pt}R{32pt} p{6pt} R{48pt}R{32pt}R{32pt}R{32pt}}
    \toprule
          & \multicolumn{4}{c}{ATE [cm] (SE3 aligned)} &                 & \multicolumn{4}{c}{RTE [cm] ($\mathrm{\Delta} = 6$ frames)}                                                                                       \\
    \cmidrule{2-5} \cmidrule{7-10}
          & ORB-SLAM3                                  & OKVIS2          & Basalt                                                       & VOCA           &  & ORB-SLAM3 & OKVIS2          & Basalt          & VOCA            \\
    \midrule
    MOO01 & \graytext{R}                               & \graytext{$\infty$}        & \graytext{$\times$}                                  & \graytext{$\times$}       &  & \graytext{R}         & \graytext{$\infty$}        & \graytext{$\times$}        & \graytext{$\times$}        \\
    MOO02 & \graytext{R}                               & \graytext{$\infty$}        & \graytext{$\times$}                                  & \graytext{$\infty$}       &  & \graytext{R}         & \graytext{$\infty$}        & \graytext{$\times$}        & \graytext{$\infty$}        \\
    MOO03 & \graytext{R}                               & \graytext{$\infty$}        & \graytext{$\infty$}                                  & \graytext{$\infty$}       &  & \graytext{R}         & \graytext{$\infty$}        & \graytext{$\infty$}        & \graytext{$\infty$}        \\
    MOO04 & \graytext{R}                               & \graytext{$\infty$}        & \graytext{$\infty$}                                  & \graytext{$\times$}       &  & \graytext{R}         & \graytext{$\infty$}        & \graytext{$\infty$}        & \graytext{$\times$}        \\
    MOO05 & R                                          & 8.05            & \textbf{1.98}                                                & 2.41           &  & R         & 2.06            & \textbf{0.52}   & 0.55            \\
    MOO06 & \graytext{R}                               & \graytext{78.98}           & \graytext{$\infty$}                                  & \graytext{\textbf{67.01}} &  & \graytext{R}         & \graytext{$\infty$}        & \graytext{$\infty$}        & \graytext{$\infty$}        \\
    MOO07 & 15.21                                      & 16.43           & 1.83                                                         & \textbf{1.60}  &  & 5.36      & 4.68            & 0.67            & \textbf{0.59}   \\
    MOO08 & \graytext{R}                               & \graytext{\textbf{80.18}}  & \graytext{$\infty$}                                  & \graytext{$\infty$}       &  & \graytext{R}         & \graytext{$\infty$}        & \graytext{$\infty$}        & \graytext{$\infty$}        \\
    MOO09 & R                                          & 1.29            & \textbf{0.80}                                                & \textbf{0.80}  &  & R         & 1.44            & \textbf{0.65}   & \textbf{0.65}   \\
    MOO10 & R                                          & 3.81            & \textbf{2.72}                                                & 2.86           &  & R         & 3.37            & 0.72            & \textbf{0.67}   \\
    MOO11 & R                                          & 37.67           & \textbf{2.66}                                                & 3.72           &  & \graytext{R}         & \graytext{$\infty$}        & \graytext{1.13}            & \graytext{\textbf{0.80}}   \\
    MOO12 & \graytext{R}                               & \graytext{$\infty$}        & \graytext{$\times$}                                  & \graytext{$\times$}       &  & \graytext{R}         & \graytext{$\infty$}        & \graytext{$\times$}        & \graytext{$\times$}        \\
    MOO13 & \graytext{R}                               & \graytext{$\infty$}        & \graytext{$\infty$}                                  & \graytext{$\infty$}       &  & \graytext{R}         & \graytext{$\infty$}        & \graytext{$\infty$}        & \graytext{$\infty$}        \\
    MOO14 & \graytext{R}                               & \graytext{$\infty$}        & \graytext{$\infty$}                                  & \graytext{$\infty$}       &  & \graytext{R}         & \graytext{$\infty$}        & \graytext{$\infty$}        & \graytext{$\infty$}        \\
    MOO15 & \graytext{R}                               & \graytext{$\infty$}        & \graytext{$\times$}                                  & \graytext{$\times$}       &  & \graytext{R}         & \graytext{\textbf{7.65}}   & \graytext{$\times$}        & \graytext{$\times$}        \\
    MOO16 & \graytext{0.61}                            & \graytext{\textbf{0.43}}   & \graytext{1.76}                                      & \graytext{$\infty$}       &  & 0.67      & 0.33            & \textbf{0.17}   & 0.31            \\
    \midrule
    AVG   & R                                          & 13.45           & \textbf{2.00}                                                & 2.28           &  & R         & 2.38            & \textbf{0.55}   & \textbf{0.55}   \\
    MED   & R                                          & 37.67           & \textbf{2.72}                                                & 67.01          &  & R         & 10.63           & 1.13            & \textbf{0.80}   \\
    SR    & 12.5\%                                     & \textbf{50.0\%} & 37.5\%                                                       & 37.5\%         &  & 12.5\%    & \textbf{37.5\%} & \textbf{37.5\%} & \textbf{37.5\%} \\
    \bottomrule
  \end{tabular}
  \label{tab:msdmo}
\end{table}

\providecommand{\graytext}[1]{\textcolor{gray}{#1}}

\begin{table}[t]\fontsize{7}{7}\selectfont
  \centering
  \caption{\textbf{\ac{MSD} - Reverb G2 MGO sequences.} The last device
    from MSD is the Reverb G2, for which, again, VOCA shows the best aggregated
    metrics and overall performance. Furthermore, VOCA is able to provide reasonable RTEs for
    two additional sequences compared to the other systems. Symbols and shading follow the conventions of \cref{tab:msdmio}.}
  \begin{tabular}{r R{48pt}R{32pt}R{32pt}R{32pt} p{6pt} R{48pt}R{32pt}R{32pt}R{32pt}}
    \toprule
          & \multicolumn{4}{c}{ATE [cm] (SE3 aligned)} &          & \multicolumn{4}{c}{RTE [cm] ($\mathrm{\Delta} = 6$ frames)}                                                                                    \\
    \cmidrule{2-5} \cmidrule{7-10}
          & ORB-SLAM3                                  & OKVIS2   & Basalt                                                       & VOCA            &  & ORB-SLAM3 & OKVIS2        & Basalt        & VOCA            \\
    \midrule
    MGO01 & \graytext{$\infty$}                        & \graytext{$\infty$} & \graytext{$\infty$}                                  & \graytext{$\infty$}        &  & \graytext{$\infty$}  & \graytext{$\infty$}      & \graytext{$\infty$}      & \graytext{$\infty$}        \\
    MGO02 & \graytext{R}                               & \graytext{$\infty$} & \graytext{$\infty$}                                  & \graytext{$\times$}        &  & \graytext{R}         & \graytext{$\infty$}      & \graytext{$\infty$}      & \graytext{$\times$}        \\
    MGO03 & \graytext{R}                               & \graytext{$\infty$} & \graytext{$\infty$}                                  & \graytext{$\infty$}        &  & \graytext{R}         & \graytext{$\infty$}      & \graytext{$\infty$}      & \graytext{$\infty$}        \\
    MGO04 & \graytext{R}                               & \graytext{$\infty$} & \graytext{$\times$}                                  & \graytext{$\infty$}        &  & \graytext{R}         & \graytext{$\infty$}      & \graytext{$\times$}      & \graytext{$\infty$}        \\
    MGO05 & R                                          & 89.05    & 3.53                                                         & \textbf{2.80}   &  & R         & 7.07          & 0.39          & \textbf{0.36}   \\
    MGO06 & \graytext{R}                               & \graytext{$\infty$} & \graytext{$\infty$}                                  & \graytext{$\infty$}        &  & \graytext{R}         & \graytext{$\infty$}      & \graytext{$\infty$}      & \graytext{\textbf{6.94}}   \\
    MGO07 & 8.45                                       & 29.18    & \textbf{3.01}                                                & 3.22            &  & $\infty$  & 4.63          & \textbf{0.44} & 0.46            \\
    MGO08 & R                                          & 91.79    & 28.42                                                        & \textbf{24.12}  &  & \graytext{R}         & \graytext{$\infty$}      & \graytext{7.24}          & \graytext{\textbf{3.16}}   \\
    MGO09 & 2.85                                       & 3.59     & 1.30                                                         & \textbf{1.26}   &  & $\infty$  & 2.44          & 0.76          & \textbf{0.74}   \\
    MGO10 & 4.36                                       & 14.07    & 1.73                                                         & \textbf{1.71}   &  & $\infty$  & 4.78          & 0.29          & \textbf{0.28}   \\
    MGO11 & 4.81                                       & 47.37    & 1.11                                                         & \textbf{1.06}   &  & \graytext{$\infty$}  & \graytext{$\infty$}      & \graytext{0.37}          & \graytext{\textbf{0.35}}   \\
    MGO12 & \graytext{R}                               & \graytext{$\infty$} & \graytext{$\infty$}                                  & \graytext{$\times$}        &  & \graytext{R}         & \graytext{$\infty$}      & \graytext{$\infty$}      & \graytext{$\times$}        \\
    MGO13 & \graytext{R}                               & \graytext{$\infty$} & \graytext{$\times$}                                  & \graytext{$\times$}        &  & \graytext{R}         & \graytext{$\infty$}      & \graytext{$\times$}      & \graytext{$\times$}        \\
    MGO14 & \graytext{$\infty$}                        & \graytext{$\infty$} & \graytext{$\infty$}                                  & \graytext{\textbf{86.32}}  &  & \graytext{$\infty$}  & \graytext{$\infty$}      & \graytext{$\infty$}      & \graytext{\textbf{9.72}}   \\
    MGO15 & \graytext{$\infty$}                        & \graytext{$\infty$} & \graytext{$\infty$}                                  & \graytext{$\infty$}        &  & \graytext{$\infty$}  & \graytext{\textbf{2.11}} & \graytext{$\infty$}      & \graytext{$\infty$}        \\
    \midrule
    AVG   & R                                          & 45.84    & 6.52                                                         & \textbf{5.69}   &  & R         & 4.73          & 0.47          & \textbf{0.46}   \\
    MED   & R                                          & 91.79    & 28.42                                                        & \textbf{24.12}  &  & R         & 11.52         & 7.24          & \textbf{3.16}   \\
    SR    & 26.7\%                                     & 40.0\%   & 40.0\%                                                       & \textbf{46.7\%} &  & 0.0\%     & 33.3\%        & 40.0\%        & \textbf{53.3\%} \\
    \bottomrule
  \end{tabular}
  \label{tab:msdmg}
\end{table}

In this section, we present detailed tables for each sequence in \ac{MSD}. We split the dataset into 4 sets. In \cref{tab:msdmio}, we show results
on the miscellaneous MIO sequences from the Valve Index, while in
\cref{tab:msdmip}, results for the highly realistic MIP playing sequences are
shown. In these sequences, the operator plays real fast-paced VR games while
recording the data. In \cref{tab:msdmo} and \cref{tab:msdmg}, we show the
sequences from the other available VR headsets, the Odyssey+ and the Reverb G2,
respectively. In most cases, VOCA shows improvement over the benchmarked
state-of-the-art systems.

\section{Comparison with DROID-SLAM}

\begin{table}[t]\fontsize{7}{7}\selectfont
    \centering
    \caption{\textbf{DROID-SLAM comparison on EuRoC.} We compare VOCA
    against DROID-SLAM at a 500k kbps bitrate. ATE is SE(3)-aligned, and RTE uses
    $\Delta=6$ frames. Lower is better for both metrics, and bold marks the
    lower valid error.}
    \begin{tabular}{r C{32pt}C{32pt} p{6pt} C{32pt}C{32pt}}
        \toprule
        & \multicolumn{2}{c}{ATE [cm] (SE3 aligned)}
        &  & \multicolumn{2}{c}{RTE [cm] ($\mathrm{\Delta} = 6$ frames)}  \\
        \cmidrule{2-3} \cmidrule{5-6}
        & DROID- & VOCA & & DROID- & VOCA \\
        & SLAM & \textbf{OURS} & & SLAM & \textbf{OURS} \\
        \midrule
        EMH01 & \textbf{6.77}  & 17.50 & & \textbf{0.50} & 0.637 \\
        EMH02 & \textbf{5.43}  & 13.30 & & \textbf{0.42} & 0.604 \\
        EMH03 & \textbf{12.18} & 20.60 & & \textbf{1.28} & 1.487 \\
        EMH04 & \textbf{20.13} & 22.90 & & \textbf{1.54} & 2.075 \\
        EMH05 & \textbf{15.41} & 20.10 & & \textbf{1.29} & 1.656 \\
        EV101 & \textbf{6.24}  & 11.30 & & 1.46 & \textbf{1.405} \\
        EV102 & \textbf{5.68}  & 8.30  & & \textbf{1.46} & 1.549 \\
        EV103 & \textbf{4.96}  & 38.30 & & \textbf{1.49} & 4.085 \\
        EV201 & 10.07 & \textbf{8.10} & & \textbf{0.47} & 2.031 \\
        EV202 & \textbf{7.56}  & 9.60  & & \textbf{0.95} & 1.446 \\
        EV203 & \textcolor{gray}{14.67} & \textcolor{gray}{$\infty^{\dagger}$}
              & & \textcolor{gray}{1.47} & \textcolor{gray}{$\infty^{\dagger}$} \\
        \midrule
        AVG & \textbf{9.92} & 17.00 & & \textbf{1.12} & 1.668 \\
        MED & \textbf{7.56} & 15.40 & & \textbf{1.29} & 1.518 \\
        \bottomrule
    \end{tabular}
    \vspace{2pt}

    \raggedright\footnotesize
    $^{\dagger}$VOCA diverged on EV203; this sequence is excluded from its
    EuRoC aggregates. DROID-SLAM aggregates use all sequences.
    \label{tab:droid-voca-euroc}
\end{table}

\begin{table}[t]\fontsize{7}{7}\selectfont
    \centering
    \caption{\textbf{DROID-SLAM comparison on TUM-VI.} We compare VOCA
    against DROID-SLAM on the TUM-VI room sequences at a 500k kbps bitrate. ATE is
    SE(3)-aligned, and RTE uses $\Delta=6$ frames. Lower is better for both
    metrics, and bold marks the lower valid error.}
    \begin{tabular}{r C{32pt}C{32pt} p{6pt} C{32pt}C{32pt}}
        \toprule
        & \multicolumn{2}{c}{ATE [cm] (SE3 aligned)}
        &  & \multicolumn{2}{c}{RTE [cm] ($\mathrm{\Delta} = 6$ frames)}  \\
        \cmidrule{2-3} \cmidrule{5-6}
        & DROID- & VOCA & & DROID- & VOCA \\
        & SLAM & \textbf{OURS} & & SLAM & \textbf{OURS} \\
        \midrule
        TR1 & 55.39 & \textbf{6.80}  & & 3.17 & \textbf{0.896} \\
        TR2 & 41.31 & \textbf{12.70} & & 8.72 & \textbf{0.941} \\
        TR3 & 59.25 & \textbf{15.90} & & 4.59 & \textbf{0.853} \\
        TR4 & 36.47 & \textbf{7.10}  & & 9.60 & \textbf{0.733} \\
        TR5 & 67.73 & \textbf{7.70}  & & 7.47 & \textbf{1.046} \\
        TR6 & 13.32 & \textbf{4.80}  & & 1.89 & \textbf{0.503} \\
        \midrule
        AVG & 45.58 & \textbf{9.20} & & 5.91 & \textbf{0.829} \\
        MED & 48.35 & \textbf{7.40} & & 6.03 & \textbf{0.874} \\
        \bottomrule
    \end{tabular}
    \label{tab:droid-voca-tumvi}
\end{table}

VOCA targets the front-end of causal visual-odometry pipelines and is therefore not designed to compete with, or replace, globally optimized SLAM systems. Rather, it is intended as a lightweight improvement to the visual tracking front-end of systems that rely on KLT-style feature tracking and that are bound to the compress-then-analyze system constraint \cite{hoferH264CompressThenAnalyzeTransmission2023}. Nevertheless, for completeness, we also report results against DROID-SLAM~\cite{teed2021droid}, a modern learning-based SLAM baseline, in \cref{tab:droid-voca-euroc} and \cref{tab:droid-voca-tumvi}.
We emphasize that this comparison should be interpreted with care. DROID-SLAM is not designed for real-time operation on low-power devices and, in its standard form, does not produce strictly causal pose estimates, since its global optimization backend refines trajectories using information from future frames. To approximate a causal setting, we disable the global backend. However, even in this configuration, a frame pose can only be produced after a subsequent keyframe has been estimated, so the method remains only approximately causal.
On EuRoC, DROID-SLAM achieves strong trajectory accuracy, as expected for a modern SLAM system. On TUM-VI, however, VOCA obtains better metrics while remaining a lightweight, GPU-free front-end method. We attribute this difference to the fisheye undistortion required for TUM-VI, which may affect DROID-SLAM more strongly than the KLT-based front-end used by VOCA.

\section{Extending VOCA to AV1}
\begin{figure}[!t]
  \centering
  \begin{subfigure}[b]{0.95\textwidth}
    \includegraphics[width=\linewidth]{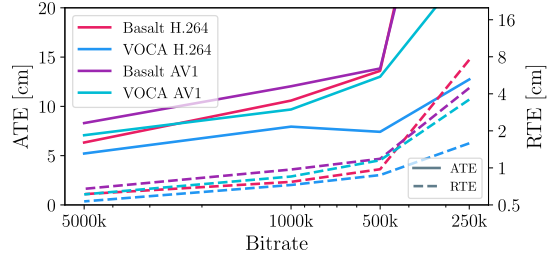}
  \end{subfigure}
  \caption{
   \textbf{H.264 and AV1 motion-vector priors on TUM-VI Room.} We show an example of decoded AV1 motion vectors and median ATE/RTE across bitrates on the TUM-VI Room sequences. Within this KLT-based setup, VOCA improves the corresponding baseline for both H.264 and AV1 priors on this dataset.
  }
  \label{fig:supp_av1}
\end{figure}

We also provide a prototypical extension of our implementation to the AV1 codec. This extension is not optimized and is intended as a proof of concept, demonstrating that the proposed approach is not tied to a specific codec and can be extended beyond H.264. We show qualitative examples of AV1 motion vectors, together with tracking results on the TUM-VI Room sequences, in \cref{fig:supp_av1}. Although preliminary, these results suggest that the improvements observed with H.264 are not specific to that codec: Within the evaluated KLT-based tracking setup, VOCA can also improve the corresponding baseline when motion-vector priors are obtained from AV1.

\immediate\closein\imgstream

\end{document}